\documentclass[letterpaper, 10 pt, conference]{ieeeconf}  

\IEEEoverridecommandlockouts                              

\usepackage{graphics} 
\usepackage{epsfig} 
\usepackage{mathptmx} 
\usepackage{times} 
\usepackage{amsmath} 
\usepackage{amssymb}  
\usepackage{booktabs}

\usepackage{multirow}
\usepackage{lipsum}
\usepackage{tcolorbox}

\usepackage{amsmath}
\usepackage{amssymb}
\usepackage[ruled,vlined]{algorithm2e}
\usepackage{graphicx}
\graphicspath{{./Figures/}}

\title{\LARGE \bf
LLM-MLFFN: Multi-Level Autonomous Driving Behavior \\ Feature Fusion via Large Language Model
}

 \author{
 	\parbox{\textwidth}{%
 		\centering
 		Xiangyu Li$^{1*}$, Tianyi Wang$^{1*}$, Xi Cheng$^{2*}$, Rakesh Chowdary Machineni$^{3}$, \\ Zhaomiao Guo$^{1\dag}$, Sikai Chen$^{4}$, Junfeng Jiao$^{5}$, Christian Claudel$^{1}$%
 	}%
\thanks{$^\dag$Corresponding author: Zhaomiao Guo.}%
\thanks{$^{*}$These authors contributed equally to this work.}%
\thanks{$^{1}$Department of Civil, Architectural, and Environmental Engineering, The University of Texas at Austin, Austin, TX 78712, USA.
 	{\tt\small xiangyu\_li@utexas.edu, bonny.wang@utexas.edu, zguo@utexas.edu, christian.claudel@utexas.edu}}%
\thanks{$^{2}$Systems Engineering Program, Cornell University, Ithaca, NY 14850, USA.
 	{\tt\small xicheng5@cornell.edu}}%
\thanks{$^{3}$Department of Electrical and Computer Engineering, University of Michigan, Ann Arbor, MI 48109, USA.
 	{\tt\small mrakeshc@umich.edu}}%
\thanks{$^{4}$Department of Civil and Environmental Engineering, University of Wisconsin-Madison, Madison, WI 53706, USA.
 	{\tt\small sikai.chen@wisc.edu}}%
\thanks{$^{5}$School of Architecture, The University of Texas at Austin, Austin, TX 78712, USA.
 	{\tt\small jjiao@austin.utexas.edu}}%
 }

\begin{document}

\maketitle
\thispagestyle{empty}
\pagestyle{empty}

\begin{abstract}

Accurate classification of autonomous vehicle (AV) driving behaviors is critical for safety validation, performance diagnosis, and traffic integration analysis.
However, existing approaches primarily rely on numerical time-series modeling and often lack semantic abstraction, limiting interpretability and robustness in complex traffic environments.
This paper presents LLM-MLFFN, a novel large language model (LLM)-enhanced multi-level feature fusion network designed to address the complexities of multi-dimensional driving data. 
The proposed LLM-MLFFN framework integrates priors from large-scale pre-trained models and employs a multi-level approach to enhance classification accuracy. 
LLM-MLFFN comprises three core components: (1) a multi-level feature extraction module that extracts statistical, behavioral, and dynamic features to capture the quantitative aspects of driving behaviors; (2) a semantic description module that leverages LLMs to transform raw data into high-level semantic features; and (3) a dual-channel multi-level feature fusion network that combines numerical and semantic features using weighted attention mechanisms to improve robustness and prediction accuracy. 
Evaluation on the Waymo open trajectory dataset demonstrates the superior performance of the proposed LLM-MLFFN, achieving a classification accuracy of over 94\%, surpassing existing machine learning models. 
Ablation studies further validate the critical contributions of multi-level fusion, feature extraction strategies, and LLM-derived semantic reasoning. 
These results suggest that integrating structured feature modeling with language-driven semantic abstraction provides a principled and interpretable pathway for robust autonomous driving behavior classification.

\end{abstract}

\section{Introduction}
\label{sec:introduction}

Over the past decade, autonomous vehicle (AV) technologies have achieved remarkable progress in perception, prediction, decision-making, and control modules \cite{yu2025bida}.
Nevertheless, robust operation in complex, mixed-traffic environments remains challenging \cite{wang2025hlcg}. 
For safe and scalable deployment, AV behaviors must be understandable and predictable to surrounding road users \cite{pan2025tgld}. 
Human-like driving can promote smoother interactions in shared environments \cite{zhang2025ccma}, but it is not a panacea: even automated systems with 360 degree perception, only about one-third of crashes can be prevented if they simply replicate human driving patterns \cite{schwarting2019social}. 
Moreover, overly conservative strategies may increase rear-end collisions and impede flow in complex scenarios such as intersections and four-way stops \cite{muhammad2020deep}. 
These observations highlight a central design tension: AV behaviors require balancing social compliance with safety and efficiency \cite{pan2025trust}.

A prerequisite for achieving this balance is the ability to systematically characterize and classify AV driving behaviors \cite{li2025svbrdllmselfverifyingbehavioralrule}. 
Prior research spans four complementary threads: semantic interaction models that encode scene context for behavior prediction \cite{hong2019rules}, conditional imitation learning that maps observations and high level commands to control \cite{codevilla2018end}, multi-modal trajectory prediction that models multiple futures with calibrated probabilities \cite{cui2019multimodal}, and cooperative perception that exploits vehicle to infrastructure data for robust situational awareness \cite{yu2023v2x}. 
An effective AV behavior classification framework should integrate these advances by aligning quantitative signals with semantic intent, and by producing predictions that are dependable and interpretable in real traffic \cite{li2026characteristics}.

Advanced machine learning and deep learning models, including random forests (RFs) \cite{breiman2001random}, convolutional neural networks (CNNs) \cite{krizhevsky2012imagenet}, long short-term memory (LSTM) networks \cite{graves2012long}, and transformer-based models \cite{vaswani2017attention}, have been widely applied in classifying driving behavior of AVs, leveraging their capabilities to process diverse and complex data for accurate behavior prediction. 
Despite considerable progress in this area, current methods exhibit notable limitations. 
Existing approaches predominantly focus on short-term trajectory forecasting or instantaneous maneuver recognition, while the broader and more stable behavioral characteristics of AV, such as aggressiveness, assertiveness, or conservativeness, remain underexplored \cite{wang2025rad}. 
Moreover, existing data-driven models often rely solely on numerical sensor signals and struggle to bridge the gap between low-level kinematic features and high-level semantic interpretations of driving intent \cite{jaeger2023hidden}. 
This limitation restricts interpretability and reduces the reliability of behavior classification in large-scale multi-modal datasets.

\begin{figure*}[htbp!]
	\centering   
    \vspace{6pt}
    \includegraphics[width=0.6\linewidth]{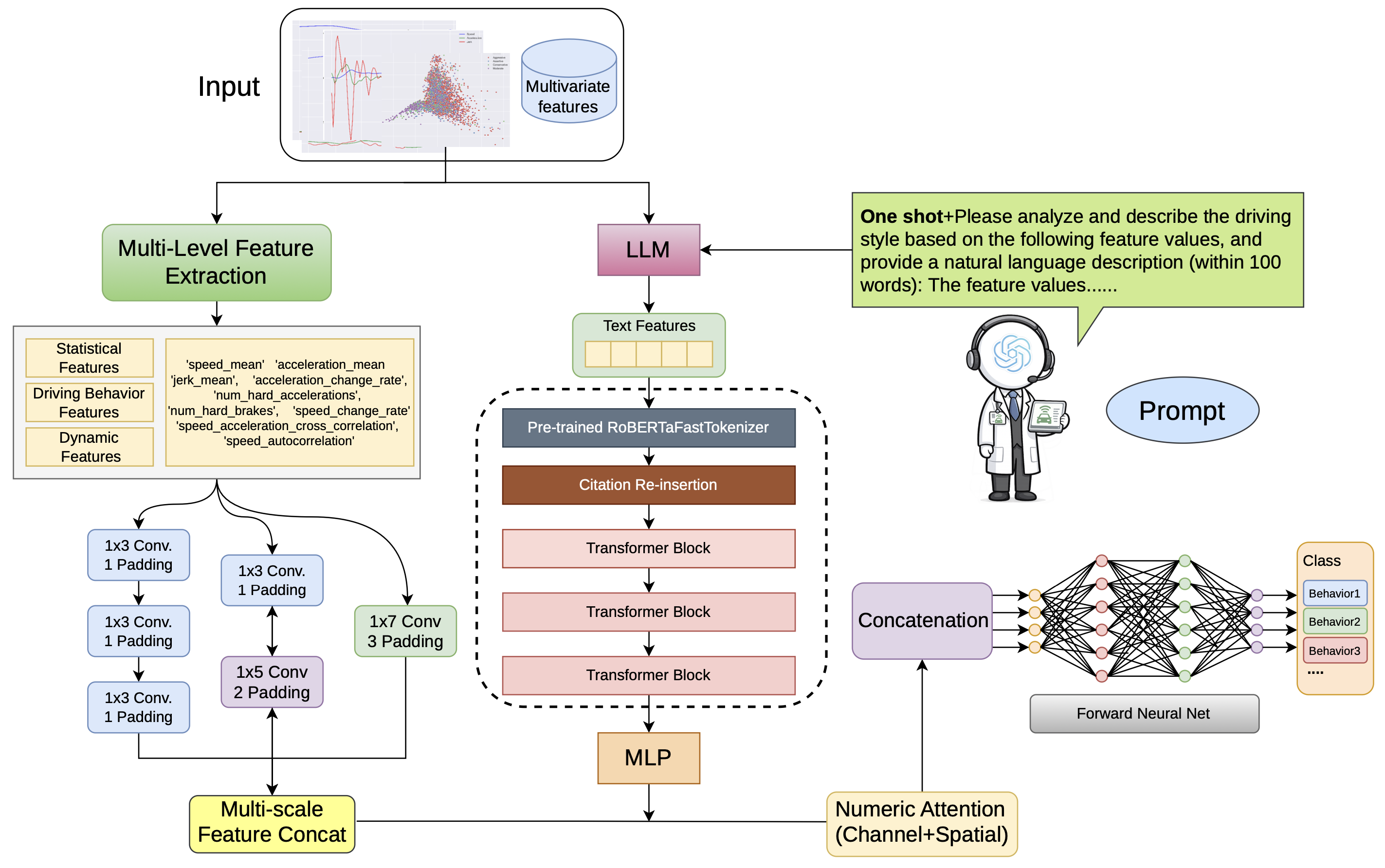}
	\caption{Overall Architecture of the Proposed LLM-MLFFN Model.}
	\label{fig:architecture}
\end{figure*}

Recent progress in large language models (LLMs) offer transformative potential in driving behavior classification \cite{wang2024rac3,wang2024drive}. 
However, directly employing LLMs as standalone classifiers for behavior prediction is unreliable due to hallucination risks, prompt sensitivity, and limited grounding in structured sensor data. 
Instead, this research focuses on feature space alignment and weighting, allowing LLMs to play a complementary role in multi-level feature fusion. 

To this end, we introduce \textbf{LLM-MLFFN} (Figure \ref{fig:architecture}), a dual-channel multi-level feature fusion framework that aligns sensor numerical features with LLM-generated semantic descriptors. 
The core idea is to establish feature-space alignment between structured statistical signals and high-level semantic representations, while preserving the reliability of supervised classification.
By assigning modality-aware weights and leveraging LLMs to map numerical patterns into interpretable semantic dimensions, the proposed \textbf{LLM-MLFFN} delivers more accurate and interpretable AV behavior classification. 
Specifically, the proposed \textbf{LLM-MLFFN} framework consists of three key modules:
\begin{itemize}
	\item \textbf{Multi-Level Feature Extraction Module:} This module extracts hierarchical quantitative descriptors from driving data, including basic statistical features (e.g., mean, standard deviation, kurtosis), driving behavior features (e.g., acceleration change rate, number of hard brakes), and dynamic features (e.g., speed-acceleration correlation). These features provide detailed quantitative insights into AV behavior.
	\item \textbf{Semantic Description Module:} This module employs prompt-guided semantic transformation to convert numerical patterns into structured textual descriptors representing behavioral intent. The generated semantic embeddings provide interpretable context and facilitate alignment between low-level signals and high-level behavior.
	\item \textbf{Dual-Channel Multi-Level Feature Fusion Network:} This module integrates statistical and semantic features using weighted attention mechanisms. The dual-channel architecture preserves numerical stability while incorporating semantic enhancement, leading to improved classification accuracy and interpretability.
\end{itemize}


\section{Related Work}
\label{sec:literature_review}

\subsection{Traditional Driving Behavior Classification}

Early work on driving behavior classification predominantly relied on single-modality inputs \cite{gopalan2012learning}. 
Vision-based methods use in-cabin cameras to detect fatigue and distraction detection via facial expressions, gaze, and head posture, or external cameras to observe lane deviations and traffic interactions \cite{ettinger2021large}. 
Although visually rich and semantically expressive, such methods are computationally intensive and exhibit performance degradation under adverse lighting and weather conditions \cite{muhammad2022vision}. 
Sensor-based methods employ accelerometers, gyroscopes, GPS, speedometers, in conjunction with machine learning models such as support vector machines (SVMs), RFs, LSTMs, or graph neural networks \cite{kumar2023driving}.
These method are resilient to visual-domain disturbances yet lack the semantic richness to characterize higher-order behavioral patterns. 
Smartphone-based alternatives have also been explored due to their portability and low deployment cost, but cross-device variability, battery limits, and privacy concerns make them ill-suited for high-precision AV tasks \cite{mantouka2021smartphone}.
Fusion strategies range from early-stage concatenation of raw inputs to late-stage merging of independently encoded representations  \cite{prakash2021multi}.
More recent work adds weighted attention mechanisms that adaptively prioritize informative features, mitigating imbalance and spatial misalignment \cite{liu2024fmdnet}. 
Multi-stream CNNs and attention-based transformers \cite{chitta2022transfuser} show strong performance in integrating vision and sensor data for tasks such as evasive-maneuver detection. 
Nevertheless, most existing frameworks process each modality independently, overlooking cross-modal semantic interactions that could further improve classification.

\subsection{Large Language Models for Autonomous Driving}

Recent breakthroughs in LLMs, such as GPT-4o \cite{achiam2023gpt}, LLaMA \cite{touvron2023llama}, and PaLM \cite{chowdhery2023palm} have introduced new paradigms for extracting high-level semantic features from both structured and unstructured data. 
LMDrive \cite{shao2024lmdrive} demonstrates closed-loop driving conditioned on natural-language goals, treating language as a supervisory signal rather than a replacement for numeric policies. DiLu \cite{wen2023dilu} wraps an LLM with explicit reasoning and reflection modules, injecting commonsense knowledge at the planning layer without intervening in low-level control.
Language has also been employed to structure and query scene semantics. 
Talk2BEV \cite{choudhary2024talk2bev} provides a visual-language interface to bird's-eye-view maps, enabling queries about layout and intent that complement kinematic statistics. 
DriveLM \cite{sima2024drivelm} contributes graph-structured language-driving tasks linking perception, prediction, and planning through explicit reasoning chains. 
DriveVLM \cite{tian2024drivevlm} shows that vision-language models improve scene description and hierarchical planning; its hybrid variant, DriveVLM-Dual, improves stability by coupling with a traditional perception stack. 
Complementary benchmarks such as nuScenes-QA \cite{qian2024nuscenes} and AutoDrive-QA \cite{khalili2025autodrive} standardize evaluation of language-aware autonomous driving systems.
Despite these advances, existing work primarily leverages LLMs for high-level goal interpretation, question answering, or planning assistance, rather than for enriching behavior representation in multi-modal classification pipelines. 
Moreover, directly relying on LLMs for label prediction risks instability due to prompt sensitivity and hallucination effects, particularly when grounding is weak.

\section{Methodology}
\label{sec:methodology}

\subsection{Multi-Level Feature Extraction Module}

This multi-level feature extraction module extracts three complementary feature sets from raw driving data.

\subsubsection{Basic Statistical Features}
For each of the $N$ raw signals $F_i(t),\; t=1,\dots,T$, we compute nine standard descriptors: mean~$\mu_i$, standard deviation~$\sigma_i$, maximum, minimum, median, 25th and 75th percentiles, kurtosis~$\kappa_i$, and skewness~$\gamma_i$. These are concatenated into:
\begin{align}
	\mathbf{F}_{\text{stat}} = [&\mu_i, \sigma_i, F_{i,\text{max}}, F_{i,\text{min}}, F_{i,\text{median}}, \nonumber \\
	&F_{i,\text{q25}}, F_{i,\text{q75}}, \kappa_i, \gamma_i]_{i=1}^{N}. \label{eq:stat_vector}
\end{align}
These features characterize global distributional properties and capture steady-state tendencies over the observation horizon.

\subsubsection{Driving Behavior Features}
To quantify maneuver intensity, we define a threshold parameter $\tau$ (set to 2 in our experiments). 
We compute the acceleration change rate~$\rho_a$, speed change rate~$\rho_v$, and counts of hard accelerations~$\ N_{\text{accel}}$, hard brakes~$\ N_{\text{brake}}$, and hard turns~$ N_{\text{turn}}$:
\begin{align}
	\rho_a &= \frac{1}{T-1} 
	\sum_{t=2}^{T} |a(t) - a(t-1)|,
	\label{eq:acc_change_rate} \\
	N_{\text{accel}} &= \sum_{t=1}^{T} 
	\mathbb{I}(a(t) > \tau), \quad
	N_{\text{brake}} = \sum_{t=1}^{T} 
	\mathbb{I}(a(t) < -\tau),
	\label{eq:num_hard_events} \\
	N_{\text{turn}} &= \sum_{t=1}^{T} 
	\mathbb{I}(|j(t)| > \tau), \quad
	\rho_v = \frac{1}{T-1} 
	\sum_{t=2}^{T} |v(t) - v(t-1)|,
	\label{eq:speed_change_rate}
\end{align}
where $\mathbb{I}(\cdot)$ is the indicator function. The resulting vector is $\mathbf{F}_{\text{behavior}} = [\rho_a,\, N_{\text{accel}},\, N_{\text{brake}},\, N_{\text{turn}},\, \rho_v]$.
These features explicitly encode volatility and maneuver aggressiveness beyond simple averages.

\subsubsection{Dynamic Features}
We capture temporal dependencies through Pearson cross-correlations between speed and acceleration ($\rho_{v,a}$) and between acceleration and jerk ($\rho_{a,j}$), together with the lag-1 autocorrelations of speed ($\rho_v^{\text{auto}}$) and acceleration ($\rho_a^{\text{auto}}$). The dynamic feature vector is 
\begin{align}
\mathbf{F}_{\text{dynamic}} = [\rho_{v,a},\, \rho_{a,j},\, \rho_v^{\text{auto}},\, \rho_a^{\text{auto}}].
\end{align}
These metrics reflect smoothness, temporal consistency, and control regularity, distinguishing conservative and aggressive driving styles.

\subsubsection{Feature Assembly and Normalization}
The three groups are concatenated into:
\begin{equation}
	\mathbf{F} = [\mathbf{F}_{\text{stat}},\, \mathbf{F}_{\text{behavior}},\, \mathbf{F}_{\text{dynamic}}] \in \mathbb{R}^D, \quad D = 9N + 5 + 4.
\end{equation}
With $N = 36$ engineered numeric features, $D = 333$. Each feature is then z-score normalized using training-set statistics before being passed to downstream modules.

\subsection{LLM Semantic Description Module}

\begin{tcolorbox}[colback=white, colframe=black, title=Prompt for Driving Behavior Analysis]
	\textbf{Example: 1 Aggressive}\\
	\textbf{Feature Values:}\\
	acceleration\_autocorrelation: 0.498655829 \\
	acceleration\_change\_rate: -0.540905602 \\
	etc...\\
	
	\textbf{User Instruction:} \\
	Please analyze the driving style based on the following feature values and describe it in natural language within 100 words.\\
	
	\textbf{LLM (i.e., GPT4o) Response:} \\
	The driver exhibits frequent and significant acceleration and deceleration, as indicated by high acceleration autocorrelation and acceleration change rate. The high jerk values and frequent occurrences of hard accelerations, brakes, and turns suggest an aggressive driving style. Additionally, the speed metrics show high and fluctuating speeds, reinforcing the characterization of this driver's style as aggressive. Overall, this driver demonstrates an aggressive driving behavior.
\end{tcolorbox}


This module uses an LLM (GPT-4o \cite{achiam2023gpt}) to convert the numerical feature vector $\mathbf{F}$ into a natural-language characterization. 
A structured prompt $P(\mathbf{F})$ lists every feature name--value pair together with a one-shot example, and the LLM returns a concise semantic analysis:
\begin{equation}
	S = \text{LLM}(P(\mathbf{F})), \label{eq:llm}
\end{equation}
where $S$ is the natural-language description of the driving behavior. The output typically covers an overall behavioral assessment, salient maneuver patterns, and explanations of key traits (e.g., high variability in acceleration change rates or frequent hard braking). Each generated description is appended to the dataset as a high-level semantic feature for downstream fusion. An example prompt is shown above.

\subsection{Dual-Channel Multi-Level Feature Fusion Network}
The fusion network merges semantic representations from the LLM module with numerical features from the extraction module through three components.

\subsubsection{Semantic Feature Channel}
A pre-trained RoBERTa-base model \cite{liu2019roberta} encodes the text $S$ into a 768-dimensional vector $\mathbf{E}$. A fully connected layer with ReLU projects it to 128 dimensions, followed by dropout regularization:
\begin{equation}
	\mathbf{E}_{\text{final}} = \text{Dropout}\!\Big(\text{ReLU}(\mathbf{W}_s \mathbf{E} + \mathbf{b}_s)\Big) \in \mathbb{R}^{128}.
\end{equation}
This branch captures high-level behavioral semantics.

\subsubsection{Numerical Feature Channel}
The numerical branch combines multi-scale convolutions, attention, and deep feature refinement:

\textbf{Multi-scale convolutions} \cite{Yu2015MultiScaleCA} with kernel sizes $k \in \{3, 5, 7\}$, each producing 64 channels, extract features at different temporal resolutions. Their outputs are concatenated into $\mathbf{C}_{\text{cat}} \in \mathbb{R}^{192 \times L}$, where $L$ is the sequence length.

\textbf{Spatio-temporal attention} highlights salient patterns:
\begin{align}
	\mathbf{Q} = \mathbf{W}_Q \mathbf{C}_{\text{cat}}, \;\;
	\mathbf{K} &= \mathbf{W}_K \mathbf{C}_{\text{cat}}, \;\;
	\mathbf{V} = \mathbf{W}_V \mathbf{C}_{\text{cat}}, \\
	\mathbf{A} &= \text{Softmax}\!\left(\frac{\mathbf{Q} \mathbf{K}^\top}{\sqrt{d_k}}\right) \mathbf{V},
\end{align}
where $\mathbf{W}_Q, \mathbf{W}_K, \mathbf{W}_V \in \mathbb{R}^{d_k \times 192}$.

\textbf{Deep refinement} follows two stacked layers of 1-D convolution, batch normalization, and ReLU activation process the attended features. A
daptive max pooling then compresses the temporal dimension, and a fully connected layer yields $\mathbf{F}_{\text{final}} \in \mathbb{R}^{128}$.

\subsubsection{Modal Fusion and Classification}
The two channel outputs are concatenated and passed through two fully connected layers to produce class logits:
\begin{align}
	\mathbf{F}_{\text{fused}} &= \text{Concat}(\mathbf{E}_{\text{final}},\, \mathbf{F}_{\text{final}}) \in \mathbb{R}^{256}, \\
	\mathbf{H} &= \text{ReLU}(\mathbf{W}_1 \mathbf{F}_{\text{fused}} + \mathbf{b}_1), \\
	\text{Logits} &= \mathbf{W}_2 \mathbf{H} + \mathbf{b}_2,
\end{align}
where $\mathbf{W}_1 \in \mathbb{R}^{256 \times 256}$ and $\mathbf{W}_2 \in \mathbb{R}^{K \times 256}$, with $K$ the number of categories of driving behavior.

\section{Experiments}
\label{sec:experiments}

\subsection{Dataset}
The proposed LLM-MLFFN model is trained and evaluated using trajectory data derived from the Waymo open dataset \cite{hu2022processing}. 
The dataset undergoes several preprocessing steps, including outlier removal and denoising, to ensure its quality and reliability. 
It captures three critical features of AV driving behavior: speed, acceleration, and jerk, which are used for both training and testing.
The dataset consists of 2,704 trips, with most trips having a duration of approximately 20 seconds and a recording interval of 0.1 seconds. 
To improve data relevance, trips where the speed remains consistently at or below zero are excluded, resulting in 2,695 meaningful trajectories for analysis. 
These trajectories form a robust foundation for understanding and classifying diverse AV driving behaviors.

\subsection{Implementation Pipeline}

The proposed LLM-MLFFN model is implemented in five key stages namely, feature extraction, semantic enhancement, feature fusion, training configuration, and model evaluation.

\subsubsection{Feature Extraction}
Feature extraction is conducted at the segment (window) level, with each CSV file representing one trajectory segment (i.e., one sample/window). The window length ~$T$ is defined as the number of time steps contained in the corresponding CSV segment. It consists:
\begin{itemize}
	\item \textbf{Basic Statistical Features:} Metrics such as mean and standard deviation are calculated to provide a quantitative summary of the dataset's central tendencies and dispersion.
	\item \textbf{Driving Behavior Features:} Features such as the number of hard accelerations and hard braking events are derived by setting specific thresholds and performing statistical analyses. These features capture explicit driving behavior patterns.
	\item \textbf{Dynamic Features:} By calculating correlations between different time points (e.g., speed-acceleration cross-correlation), these features capture the temporal dependencies and dynamic changes in driving behavior.
\end{itemize}

\subsubsection{Semantic Enhancement}
The semantic enhancement process leverages the natural language processing capabilities of LLMs to convert traditional numerical features into high-level semantic descriptions. Specific prompts are designed to generate semantic information that captures the context and patterns underlying driving behavior. For example, a prompt might generate a description like, “The driver exhibits frequent hard braking, indicating an aggressive driving style.” This transformation enhances the model's understanding of autonomous driving behaviors.

\subsubsection{Feature Fusion}
In the feature fusion stage, we obtain a semantic embedding from RoBERTa and a numerical embedding from the multi-scale convolution (with intra-branch spatio-temporal attention). We then fuse the two modality-specific embeddings via concatenation and feed the fused representation into a multi-layer perceptron (MLP) classifier.

\begin{figure}[htbp!]
	\centering
    \vspace{6pt}
	\includegraphics[width=0.5\textwidth]{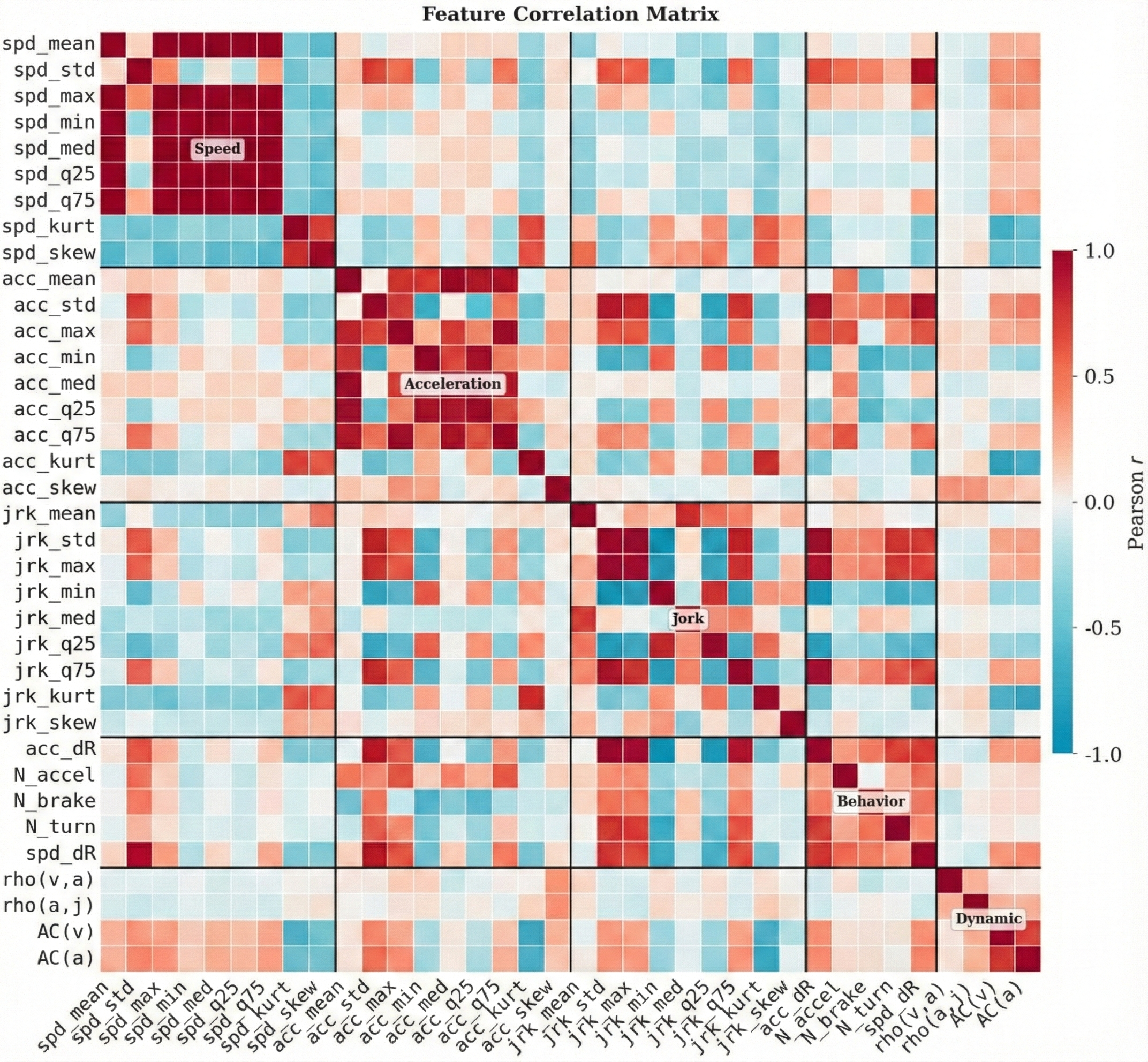}
	\caption{Feature Correlation Matrix.}
	\label{fig:feature_correlation_matrix}
\end{figure}

\subsubsection{Training Configuration}
The LLM-MLFFN model is trained in a supervised manner, where input feature vectors and their corresponding labels are used to compute the loss and update model parameters via backpropagation. For the text encoder, inputs are tokenized with a maximum sequence length of 256. We optimize the model using the cross-entropy loss for multi-class classification. To mitigate overfitting, we apply Dropout (rate $0.3$) and $L_2$ regularization. We use the AdamW optimizer with a learning rate of $2\times10^{-5}$ and a batch size of 64 to enable stable and efficient gradient updates, leading to improved convergence behavior. For the LLM-based description generation, we use a fixed one-shot prompt and set the decoding temperature to 0.5 with a maximum output length of 500 tokens.

\subsubsection{Model Evaluation}
After training, the model is evaluated using metrics such as accuracy, precision, recall, and F1-score. Cross-validation is applied to ensure objective evaluation, with the dataset randomly split into 80\% for training, 10\% for validation, and 10\% for testing. Each metric is computed as:
\begin{align}
	\text{Precision} &= \frac{TP}{TP + FP}, \\
	\text{Recall} &= \frac{TP}{TP + FN}, \\
	\text{F1-Score} &= 2 \cdot \frac{\text{Precision} \cdot \text{Recall}}{\text{Precision} + \text{Recall}}, \\
	\text{Accuracy} &= \frac{TP + TN}{TP + TN + FP + FN},
\end{align}
where \( TP \), \( TN \), \( FP \), and \( FN \) represent the number of true positive, true negative, false positive, and false negative, respectively.

This comprehensive pipeline ensures a robust and systematic approach to training and evaluating the LLM-MLFFN model, achieving high reliability in driving behavior classification tasks.

\begin{figure}[htbp!]
	\centering
    \vspace{6pt}
	\includegraphics[width=0.5\textwidth]
    {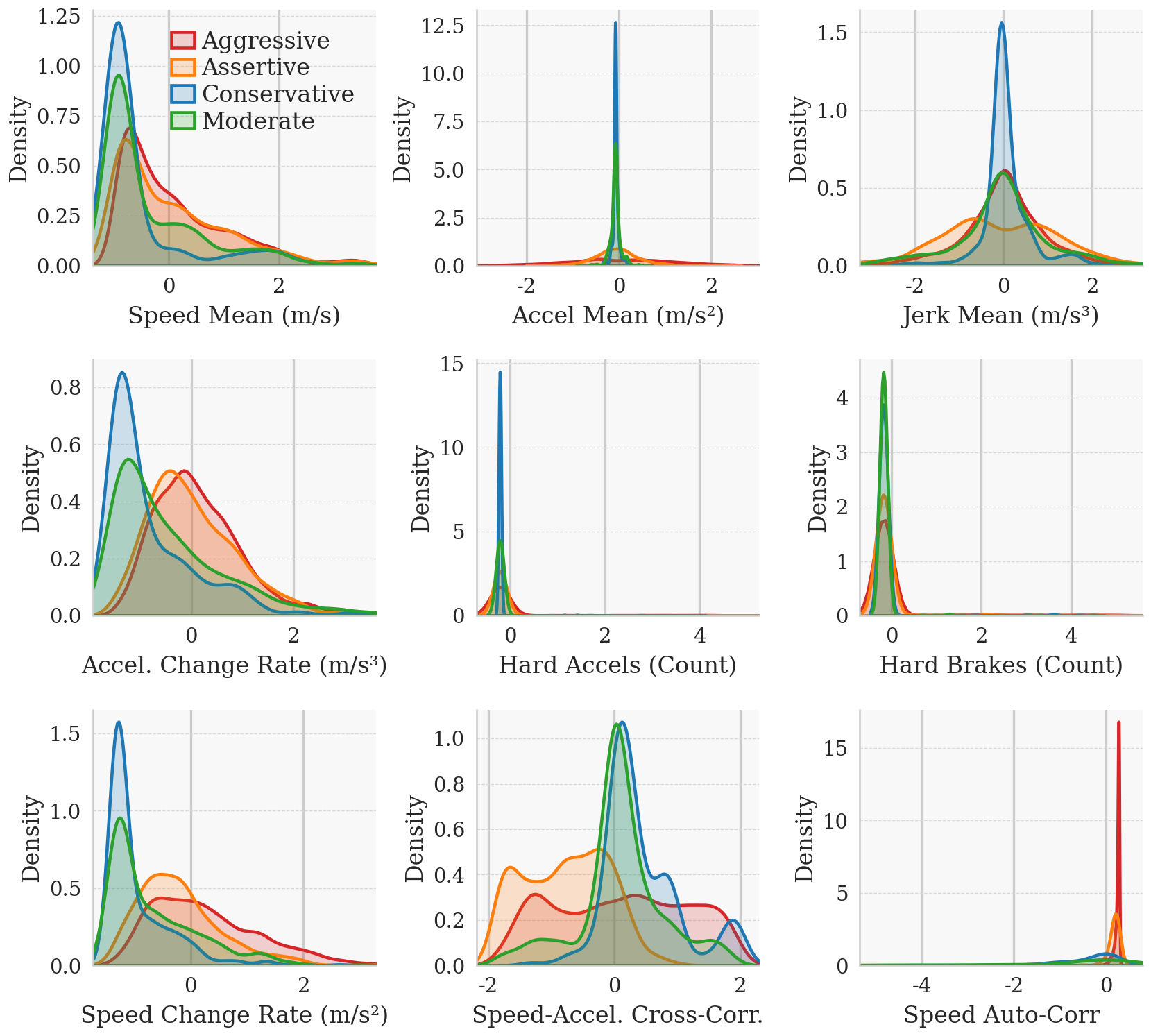}
    
	\caption{Feature Distributions for Different Driving Behavior Types (Aggressive, Assertive, Conservative, and Moderate).}
	\label{fig:feature_distribution}
\end{figure}

\subsection{Feature Analysis}

To reduce dimensionality, we first filter redundant variables using a feature correlation matrix (Figure~\ref{fig:feature_correlation_matrix}). 
The distributions of the nine selected representative features across the four driving styles (Aggressive, Assertive, Conservative, and Moderate) are then examined via kernel density estimates (Figure~\ref{fig:feature_distribution}), yielding the critical insights detailed below.

For the $speed\_mean$ feature, the distributions of Aggressive and Assertive driving styles are skew toward higher average speeds, indicating a preference for faster driving. Notably, the Aggressive style exhibits significantly higher speed averages compared to other categories. In contrast, Conservative and Moderate styles are concentrated in lower-speed regions, consistent with the cautious nature of Conservative drivers. This demonstrates that $speed\_mean$ effectively distinguishes between aggressive and conservative driving behaviors, particularly excelling in identifying Aggressive driving.

The $acceleration\_mean$ feature shows substantial overlap across all driving styles, especially near zero. This overlap suggests that limited discriminatory power for distinguishing between driving styles, as the mean acceleration appears balanced across all categories, failing to provide sufficient differentiation.

For $jerk\_mean$ (mean rate of change of acceleration), the distributions for Conservative and Moderate driving styles are more concentrated, whereas Aggressive and Assertive styles are more dispersed. This indicates that $jerk\_mean$ effectively captures the smoothness of driving behavior. Conservative drivers exhibit lower jerk values, reflecting smoother driving patterns, while Aggressive drivers show higher jerk values, indicating more abrupt acceleration and deceleration. Thus, jerk\_mean is sensitive in differentiating aggressive and conservative driving styles, capturing the volatility of Aggressive driving.

For $acceleration\_change\_rate$, Aggressive and Assertive styles are concentrated in regions of higher change rates, reflecting more frequent and intense acceleration changes. Conservative styles, on the other hand, are associated with lower change rates, indicating steadier driving patterns. This feature is highly effective in distinguishing aggressive from conservative driving behaviors, particularly highlighting the high variability characteristic of Aggressive driving.

The $num\_hard\_accelerations$ feature provides a clear distinction among driving styles. Aggressive driving exhibits significantly higher counts of hard accelerations compared to other styles, while Conservative and Moderate styles are concentrated in regions with fewer hard accelerations. This indicates that $num\_hard\_accelerations$ is a strong feature for identifying aggressive driving behavior, providing intuitive insights into acceleration habits.

For the $num\_hard\_brakes$ feature, Aggressive driving styles show notably higher counts of hard braking events, reflecting frequent and abrupt deceleration maneuvers. Conservative and Moderate styles exhibit minimal hard braking, indicating smoother and more stable driving patterns. Consequently, $num\_hard\_brakes$ emerges as a key feature for identifying aggressive driving behaviors.

The $speed\_change\_rate$ feature is more concentrated in negative regions for Aggressive driving, suggesting frequent deceleration. In contrast, Moderate and Conservative styles are closer to zero, indicating less variability in speed. This highlights the significant fluctuations in speed characteristic of Aggressive driving, making $speed\_change\_rate$ a strong feature for differentiating this style from others.

Regarding $speed\_acceleration\_cross\_correlation$, Conservative and Moderate styles exhibit more concentrated distributions in regions of higher correlation, indicating smoother relationships between speed and acceleration. Aggressive and Assertive styles, on the other hand, show more dispersed distributions, reflecting greater instability in their acceleration-deceleration patterns. This feature is particularly effective in identifying Conservative driving styles, especially in distinguishing between smooth and aggressive behaviors.

Lastly, the $speed\_autocorrelation$ feature reveals concentrated distributions near zero, with Conservative and Moderate styles showing greater concentration in higher autocorrelation regions. This indicates that the feature captures the regularity of driving patterns, particularly reflecting the smoothness of Conservative driving. However, its contribution to distinguishing different driving styles is limited and serves primarily as a supplementary feature.

In conclusion, the features $num\_hard\_accelerations$, $num\_hard\_brakes$, $acceleration\_change\_rate$, and $speed\_change\_rate$ are the most effective for differentiating driving styles, particularly between Aggressive and Conservative driving. These features effectively capture acceleration, braking, and speed variation patterns, serving as critical indicators for identifying distinct driving behaviors. While $speed\_mean$ and $jerk\_mean$ provide supplementary information, their overlapping distributions limit their overall contribution to classification.



\subsection{Comparative Experiment}
To rigorously evaluate the proposed LLM-MLFFN framework, we benchmark its performance against nine established multivariate time series classification models. These baselines are selected to represent three distinct architectural paradigms:

\begin{itemize}
    \item \textbf{Standard Deep Learning:} We include foundational models such as the (\textbf{MLP}) \cite{wang2017time}, as well as \textbf{LSTM} \cite{graves2012long} and fully convolutional networks (\textbf{FCN}) \cite{wang2017time}, which serve as standard benchmarks for capturing standalone temporal and spatial features, respectively.
    \item \textbf{Hybrid and Multi-Scale Networks:} To evaluate combinations of recurrent and convolutional feature extraction, we compare against \textbf{LSTM-FCN} \cite{karim2017lstm} and \textbf{GRU-FCN} \cite{elsayed2018deep}. We also include multi-scale architectures designed to capture local and global patterns simultaneously, specifically \textbf{mWDN} \cite{wang2018multilevel} and \textbf{MLSTM-FCN} \cite{karim2019multivariate}.
    \item \textbf{Transformer-Based Architectures:} Finally, we benchmark against recent attention-based state-of-the-art models. This includes the time series transformer (\textbf{TST}) \cite{zerveas2021transformer} for temporal self-attention, and \textbf{GAF-ViT} \cite{you2024exploring}, which leverages vision transformers to process time-series data converted into visual graph structures.
\end{itemize}

\begin{table*}[htbp]
	 \vspace{6pt}
  \caption{Comparison of Feature-Engineered and Non-Feature-Engineered Models Across Different Metrics}
	\begin{center}
		\resizebox{0.8\linewidth}{!}{
		\begin{tabular}{ccccccccc}
			\hline
			\multirow{2}{*}{\textbf{Model}} & \multicolumn{2}{c}{\textbf{Acc.$\uparrow$}} & \multicolumn{2}{c}{\textbf{Pre.$\uparrow$}} & \multicolumn{2}{c}{\textbf{Rec.$\uparrow$}} & \multicolumn{2}{c}{\textbf{F1$\uparrow$}} \\
			\cline{2-9}
			& \textbf{Non-Feat.} & \textbf{Feat.} & \textbf{Non-Feat.} & \textbf{Feat.} & \textbf{Non-Feat.} & \textbf{Feat.} & \textbf{Non-Feat.} & \textbf{Feat.} \\
			\hline
			LSTM \cite{graves2012long}        & 0.7166 & 0.8888 & 0.6227 & 0.8925 & 0.4836 & 0.8888 & 0.4955 & 0.8895 \\
			MLP \cite{wang2017time}         & 0.8321 & 0.8824 & 0.8584 & 0.8829 & 0.6721 & 0.8824 & 0.7394 & 0.8812 \\
			FCN \cite{wang2017time}         & 0.8075 & 0.7519 & 0.7915 & 0.7615 & 0.6540 & 0.7519 & 0.7040 & 0.6943 \\
			LSTM-FCN \cite{karim2017lstm}    & 0.8032 & 0.8909 & 0.8080 & 0.8981 & 0.6334 & \underline{0.8909} & 0.6940 & \underline{0.8934} \\
			GRU-FCN \cite{elsayed2018deep}     & 0.6909 & 0.8877 & 0.5536 & \underline{0.8955} & 0.4554 & 0.8877 & 0.4782 & 0.8893 \\
			mWDN \cite{wang2018multilevel}        & 0.9005 & 0.8684 & 0.8595 & 0.8801 & 0.8224 & 0.8684 & 0.8385 & 0.8703 \\
			MLSTM-FCN \cite{karim2019multivariate}   & 0.8182 & 0.8299 & 0.8003 & 0.8409 & 0.6843 & 0.8299 & 0.7311 & 0.8140 \\
			TST \cite{zerveas2021transformer}         & 0.7508 & 0.7701 & 0.7896 & 0.7622 & 0.4985 & 0.7701 & 0.5586 & 0.7347 \\
			GAF-ViT \cite{you2024exploring}     & \textbf{0.9209} & \underline{0.9219} & \underline{0.8679} & 0.8800 & \underline{0.8826} & 0.8900 & \underline{0.8747} & 0.8850 \\
			\textbf{LLM-MLFFN (Ours)}    & \underline{0.9145} & \textbf{0.9430} & \textbf{0.9158} & \textbf{0.9464} & \textbf{0.9145} & \textbf{0.9430} & \textbf{0.9135} & \textbf{0.9414} \\
			\hline
		\end{tabular}}
		\label{tab:comparison_metrics}
	\end{center}
\end{table*}

\paragraph{Feature-Engineered vs. Non-Feature-Engineered Models} Experiments are conducted to compare the performance of feature-engineered models (utilizing the proposed multi-level feature extraction module) with non-feature-engineered models. 
The results, presented in Table~\ref{tab:comparison_metrics}, demonstrate that feature-engineered models significantly outperform their non-feature-engineered counterparts across all evaluation metrics. 
This highlights the effectiveness of the proposed feature extraction module.

\paragraph{Performance Comparison}
The experimental results shown in Table~\ref{tab:comparison_metrics} indicate that the proposed LLM-MLFFN model consistently outperforms all baseline models across all metrics. 
Notably, our LLM-MLFFN model demonstrates superior performance in both feature-engineered and non-feature-engineered settings, highlighting its robust multi-level feature extraction and fusion capabilities. 
While models such as GAF-ViT \cite{you2024exploring}, LSTM-FCN \cite{karim2017lstm} and mWDN \cite{wang2018multilevel} also achieve strong results, LLM-MLFFN surpasses them in precision and stability across various time series tasks, due to its advanced deep feature extraction and semantic modeling capabilities.



\subsection{Ablation Study}

To assess the contribution of each component in the LLM-MLFFN model, an ablation study is conducted by systematically removing or replacing key modules within the framework. 
The results, presented in Table~\ref{tab:ablation_study}, highlight the significance of each component in the model's overall performance.

\begin{table}[htbp]
    \vspace{6pt}
    \caption{Ablation Study Results}
    \begin{center}
        \footnotesize 
        \begin{tabular*}{\columnwidth}{@{\extracolsep{\fill}}lcccc@{}}
            \hline
            \textbf{Model Variation} & \textbf{Acc.$\uparrow$} & \textbf{Pre.$\uparrow$} & \textbf{Rec.$\uparrow$} & \textbf{F1$\uparrow$} \\
            \hline
            \textbf{Full Model} & \textbf{0.9430} & \textbf{0.9464} & \textbf{0.9430} & \textbf{0.9414} \\
            w/o Spatio-Temp Attn. & 0.9311 & 0.9333 & 0.9311 & 0.9298 \\
            w/o Multi-Scale Conv. & 0.9359 & 0.9409 & 0.9359 & 0.9343 \\
            Text Features Only & 0.9145 & 0.9158 & 0.9145 & 0.9135 \\
            Num. Features Only & 0.9144 & 0.9161 & 0.9144 & 0.9147 \\
            \hline
        \end{tabular*}
        \label{tab:ablation_study}
    \end{center}
\end{table}

The removal of the spatio-temporal attention mechanism results in a notable decline in performance, with reductions across all evaluation metrics, including accuracy, precision, recall, and F1-score. \textit{Note that the reported recall equals accuracy because we report micro-averaged recall in a single-label multi-class setting, where both metrics reduce to the fraction of correctly classified samples.} This demonstrates the critical role of the spatio-temporal attention mechanism in capturing the temporal and spatial dependencies within the data.
Similarly, excluding the multi-scale convolution module leads to a further decline in performance, particularly in precision and F1-score. These findings emphasize the importance of multi-scale convolutions in extracting multi-level features and enhancing the model’s ability to discern nuanced driving behaviors.
When the model is evaluated using only semantic features or numerical features individually, its classification performance is significantly lower than that of the complete model. This is especially evident in precision and recall, highlighting the necessity of fusing numerical and semantic features. The integration of these feature types enables the model to leverage the strengths of both modalities, achieving a more comprehensive representation of driving behaviors.


\section{Conclusion}
\label{sec:conclusion}

This study proposes \textbf{LLM-MLFFN} (Large Language Model-Enhanced Multi-Level Feature Fusion Network), a novel framework for autonomous driving behavior analysis, designed to comprehensively classify and understand driving behaviors by integrating statistical, behavioral, and dynamic features. 
The proposed \textbf{LLM-MLFFN} framework incorporates a multi-level feature extraction module, a semantic description module, and a dual-channel multi-level feature fusion network, significantly enhancing the accuracy and performance of autonomous driving behavior classification.
Experimental results on the Waymo open dataset demonstrate that our \textbf{LLM-MLFFN} model outperforms existing methods across multiple evaluation metrics, highlighting its ability to capture the complexity of autonomous vehicle behaviors. 
Ablation studies validate the critical role of the multi-level feature extraction and semantic description modules, while confirming the effectiveness of the dual-channel architecture in leveraging the complementary strengths of numerical and semantic features. 
By seamlessly combining numerical and textual features, the proposed framework provides a holistic approach to analyze complex driving behaviors.
Future research directions include extending the proposed method to more complex traffic environments and optimizing its real-time processing capabilities and computational efficiency. Additionally, incorporating diverse sensor data and exploring more varied driving scenarios will enhance the model's robustness and generalization. These efforts will support the applicability and scalability of the \textbf{LLM-MLFFN} framework, ensuring its relevance to real-world autonomous driving systems.


\bibliographystyle{IEEEtran}
\bibliography{references}

\end{document}